\relax
\documentclass[letterpaper]{article} 
\usepackage{aaai18_mod}  
\usepackage{times}  
\usepackage{helvet}  
\usepackage{courier}  
\usepackage{url}  
\usepackage{graphicx}  

\usepackage{epsfig}
\usepackage{graphicx}
\usepackage{amsmath}
\usepackage{amssymb}
\usepackage{array}

\begin{document}
%
\title{Extreme Low Resolution Activity Recognition with Multi-Siamese \\Embedding Learning}

\author{Michael S. Ryoo$^{1,2}$, Kiyoon Kim$^{1,3}$, Hyun Jong Yang$^{1,3}$\\
$^1$EgoVid Inc., Daejeon, South Korea\\
$^2$Indiana University, Bloomington, IN, USA\\
$^3$Ulsan National Institute of Science and Technology, Ulsan, South Korea\\
\texttt{\{mryoo, hjyang\}@egovid.com}\\
}

\maketitle

\begin{abstract}
This paper presents an approach for recognizing human activities from \emph{extreme low resolution} (e.g., 16x12) videos. Extreme low resolution recognition is not only necessary for analyzing actions at a distance but also is crucial for enabling privacy-preserving recognition of human activities. We design a new two-stream multi-Siamese convolutional neural network. The idea is to explicitly capture the inherent property of low resolution (LR) videos that two images originated from the exact same scene often have totally different pixel values depending on their LR transformations. Our approach learns the shared embedding space that maps LR videos with the same content to the same location regardless of their transformations. We experimentally confirm that our approach of jointly learning such transform robust LR video representation and the classifier outperforms the previous state-of-the-art low resolution recognition approaches on two public standard datasets by a meaningful margin.
\end{abstract}



\section{Introduction}

Although there has been a large amount of progress in human activity recognition research in the past years \cite{ryoo-review,simonyan14,google15,c3d}, most of the existing works assume that region-of-interest (ROI) in videos are large enough. The assumption is that each video region corresponding to an activity has a high enough resolution, allowing the recognition model to capture detailed motion and appearance changes. However, there are several cases where this assumption does not hold. For instance, in far-field recognition scenarios (i.e., detecting human activities at a distance), humans are usually very far away from the camera and each ROI often has just a few pixels within. This happens commonly in visual surveillance cameras \cite{efros2003recognizing,reddy2012human}, required to cover a large area while having a low native resolution due to their cost.


\begin{figure}
\begin{center}
   \includegraphics[width=1.0\linewidth]{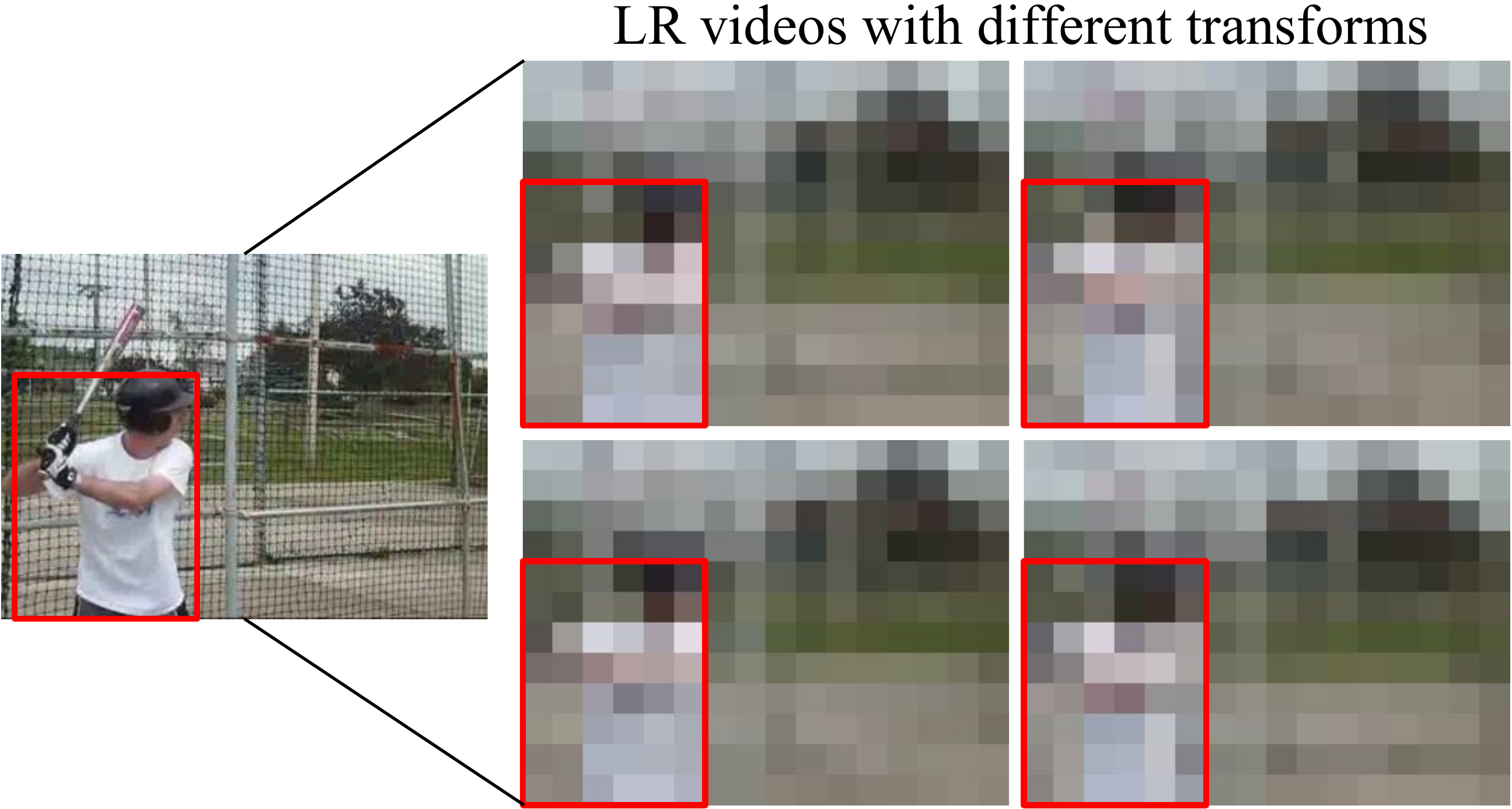}
   \includegraphics[width=1.0\linewidth]{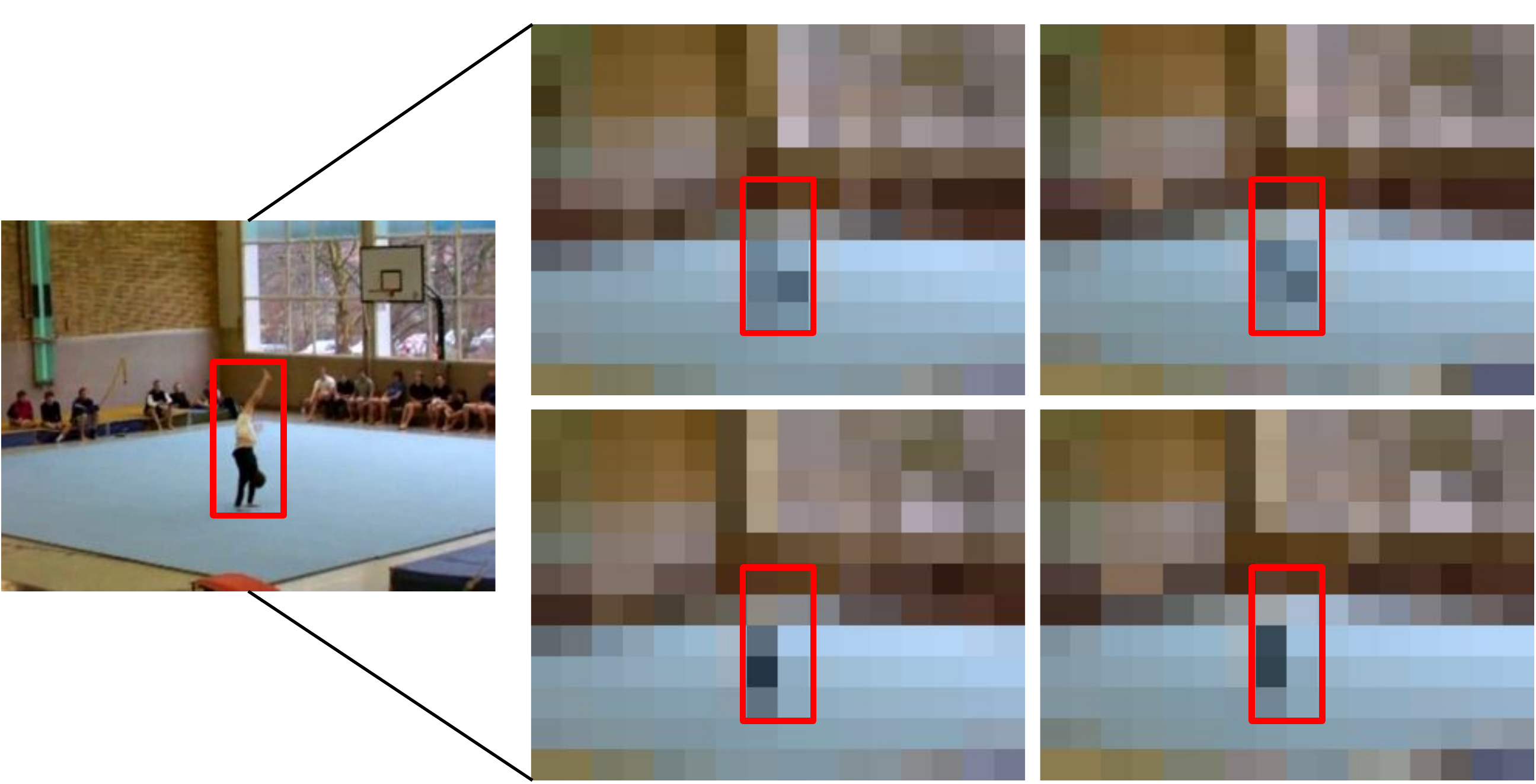}
\end{center}
   \caption{Example LR images generated by applying different LR transforms (with slightly different translations) to a single HR image. Red boxes indicate pixels of the humans. Although these LR images (right) are all from the identical HR frame (left), their pixel values become very different.}
\label{fig:intro}		
\end{figure}

Furthermore, there are situations where one wants to intentionally avoid taking high-resolution (HR) videos because of a privacy concern. High resolution cameras including robot cameras and wearable cameras are becoming increasingly available at both public and private places, and we are afraid of them recording privacy-sensitive videos of us without consent. For example, if such camera system at home (for home security or smart home services) is cracked by a hacker, there is a risk of one's 24/7 private life being monitored/recorded by someone else. The paradigm of using extreme low resolution (e.g., 16x12) anonymized videos for \emph{privacy-preserving} activity recognition is able to address such societal concern of unwanted video taking at the fundamental-level. Human faces in extreme LR videos are not identifiable (e.g., they are much smaller than 5x5), naturally prohibiting the recognition process from accessing privacy-sensitive face information. This allows designing the device (e.g., a robot) that does not record HR videos while still recognizing what is going on around it for its operation. Although extreme low resolution videos are not the only privacy-preserving data (e.g., super-pixeled frames could also be privacy-preserving), they probably are the most computation (and hardware) efficient data to obtain/process and a number of recent research \cite{dai15,ryoo17privacy} studied such direction.

Motivated by such demands, there were several previous studies on extreme low resolution object/activity recognition \cite{dai15,lrface16,ryoo17privacy,chen17,cheng17emotion}. The learning in previous approaches was typically done by resizing each original high resolution training sample to a LR sample and using it as a training data. On the other hand, although the recognition methods are required to only use extreme low resolution data in the testing phase, it is a realistic assumption to use publicly available HR data (e.g., YouTube videos) for their learning in the training phase. Several previous works took such direction/assumption \cite{lrface16,ryoo17privacy,chen17,cheng17emotion} for the better LR recognition and obtained promising results.

However, most of the previous works were limited in the aspect that they seldom considered the intrinsic property of low resolution sensors: In LR images, due to the inherent limitation what a single pixel can capture from the scene, two images originated from the exact same scene often have totally different pixel (i.e., RGB) values. Camera transformations (particularly motion transformations \cite{huang10}) such as sub-pixel translations and rotations influence the image data significantly. Figure \ref{fig:intro} shows an example. Depending on the transformations, LR images from the exact same scene become different visual data.

In this paper, we propose a new low resolution classification approach that explicitly takes such property into account to enable better recognition of human activities from LR videos. The idea is that multiple LR videos (e.g., Figure \ref{fig:intro}) corresponds to a single HR video and thus should ideally be embedded to the same representation (to be used for the classification). That is, the intermediate representations corresponding to these LR videos should be very similar, mapping the videos to the same point in the embedding space. 
Once such embedding space is jointly learned with its classifier, when a new LR video is provided in the testing phase, the model can map the video to its corresponding embedding location regardless of its (unknown) LR transform. This means that the classifier becomes invariant to sub-pixel transforms (e.g., affine transforms including translation, scaling, and rotation) of the LR camera. A new multi-Siamese Convolutional Neural Network (CNN) architecture is designed to learn the optimal embeddings for LR videos.



We experimentally confirm that our concept of posing an additional constraint in the representation (i.e., embedding) learning that ``LR videos corresponding to the same HR videos should be identical/similar'' obtains better performance than the conventional approach of learning a classifier with the exact same number of augmented LR training videos. Our approach jointly optimizes the video representation and the classifier for the best LR activity recognition, obtaining superior performances to prior works.

\section{Related works}

Human activity recognition is an important research area actively studied since 1990s \cite{ryoo-review}. In the past 3 years, approaches taking advantage of video-based convolutional neural networks showed particularly successful results in activity recognition. These not only include the approaches to capture relatively short-term (e.g., 15 frames) motion in videos such as two-stream CNN \cite{simonyan14} and C3D \cite{c3d}, but also include those to capture longer-term temporal structure like long-term temporal convolution \cite{varol16} and temporal attention filters \cite{piergiovanni2016learning}. Use of recurrent neural networks (RNNs) to model sequential changes in activity videos also have been popular \cite{google15,yeong16}. The approaches obtained successful results particularly in video classification. However, they did not consider activity recognition from low resolution videos (their target resolution was at least 200x200) and thus was not suitable for LR recognition as they are.

There have been more recent works on extreme low resolution activity recognition \cite{dai15,ryoo17privacy,chen17,cheng17emotion}. Some of these works focused on obtaining better low resolution features \cite{dai15}. Other works focused on taking advantage of high resolution training videos to learn better LR decision boundaries. The idea was that one high-resolution training image/video contains more information than just a single low-resolution data.
\cite{ryoo17privacy} considered that multiple different LR transforms can be used to increase the number of training data from a single HR video, although it did not attempt any LR representation learning. \cite{chen17} took advantage of the LR face recognition approach introduced in \cite{lrface16}; they designed the video version of \cite{lrface16}. Features to be shared in both HR and LR videos were learned in this approach. However, it did not take advantage of the fact that there can be multiple LR transforms, and its recognition accuracy was thus limited.


There were previous works on Siamese CNNs for various different computer vision problems (e.g., \cite{hadsell2006dimensionality,bell15siggraph,wang2015unsupervised}), but we believe this is the first paper to conduct the Siamese embedding learning for low resolution data. Previous Siamese CNNs were not focusing on exploiting the properties of LR data, and we are not aware of any such  attempts for LR videos or activity recognition. Our approach is also different from the general data augmentation method increasing the number of training data; our method explicitly learns the intermediate LR embedding while considering sub-pixel transformations in LR videos, thereby becoming transform robust and performing superior.



\section{Our approach}

In this section, we describe our approach to recognize human activities from extreme low resolution videos. The key idea is that (1) multiple different LR transforms can be applied to a single HR training video to obtain a set of LR videos and that (2) we can learn the `embedding space' that explicitly forces intermediate CNN representations of such LR videos to be transform invariant while jointly optimizing them for the classification. We assume the availability of HR training videos from publicly available sources (e.g., YouTube), and present a method to best take advantage of such HR training videos to learn the optimal LR classifier.


Given a set of original HR training videos, the goal is to learn the \textbf{embedding space for low resolution videos}, and use the learned embedding for the classification of a new LR testing video. The learned embedding ideally maps LR videos (from the same original HR video) to the same location regardless of their transformations, thereby enabling learning of transform-invariant activity classifiers. Rather than using a hand designed mapping, we use a Siamese CNN architecture while explicitly designing it to handle multiple LR transforms. A two-stream network for extreme LR videos is presented, and a new Siamese architecture with multiple branches for the extreme LR classification (i.e., our multi-Siamese CNN architecture) is introduced.


\subsection{Low resolution video transforms}
\label{subsec:lr}



Motivated by the finding that the use of multiple different LR transforms benefits the classifier learning \cite{ryoo17privacy}, we designed our approach to explicitly take advantage of a set of LR transforms. The main idea is that a single high resolution video contains an equivalent amount of information to a set of low resolution videos, and the recognition approaches can exploit that by applying different LR transforms to a single HR video. We generate $n$ number of low resolution videos (i.e., $V_{ik}$) for each high resolution training video $X_i$ by applying the set of transforms $F_k$ and $D_k$:
\begin{equation}
    V_{ik} = D_k F_k X_i, ~~k = 1 \ldots n.
\end{equation}
where $F_k$ is the camera motion transformation and $D_k$ is the down-sampling operator. Here, $F_k$ can be any affine transformation, but we consider combinations of translation, scaling, and rotation as our motion transform in this paper. We use the standard average downsampling for $D_k$.

Unlike \cite{ryoo17privacy} which attempted to learn a smaller subset of transforms computationally efficient for the training of the classifiers, in this paper, we take the strategy of providing a sufficient number of transforms to the classifier,  $S = \{F_k\}^n_{k=1}$, and attempt to best take advantage of them to maximize the classification performance. Multiple $V_{ik}$ generated from each training sample $X_i$ will be used for the training of our approach, which we present in the ``Multi-Siamese CNN'' subsection in more detail.

\begin{figure}
\begin{center}
   \includegraphics[width=0.8\linewidth]{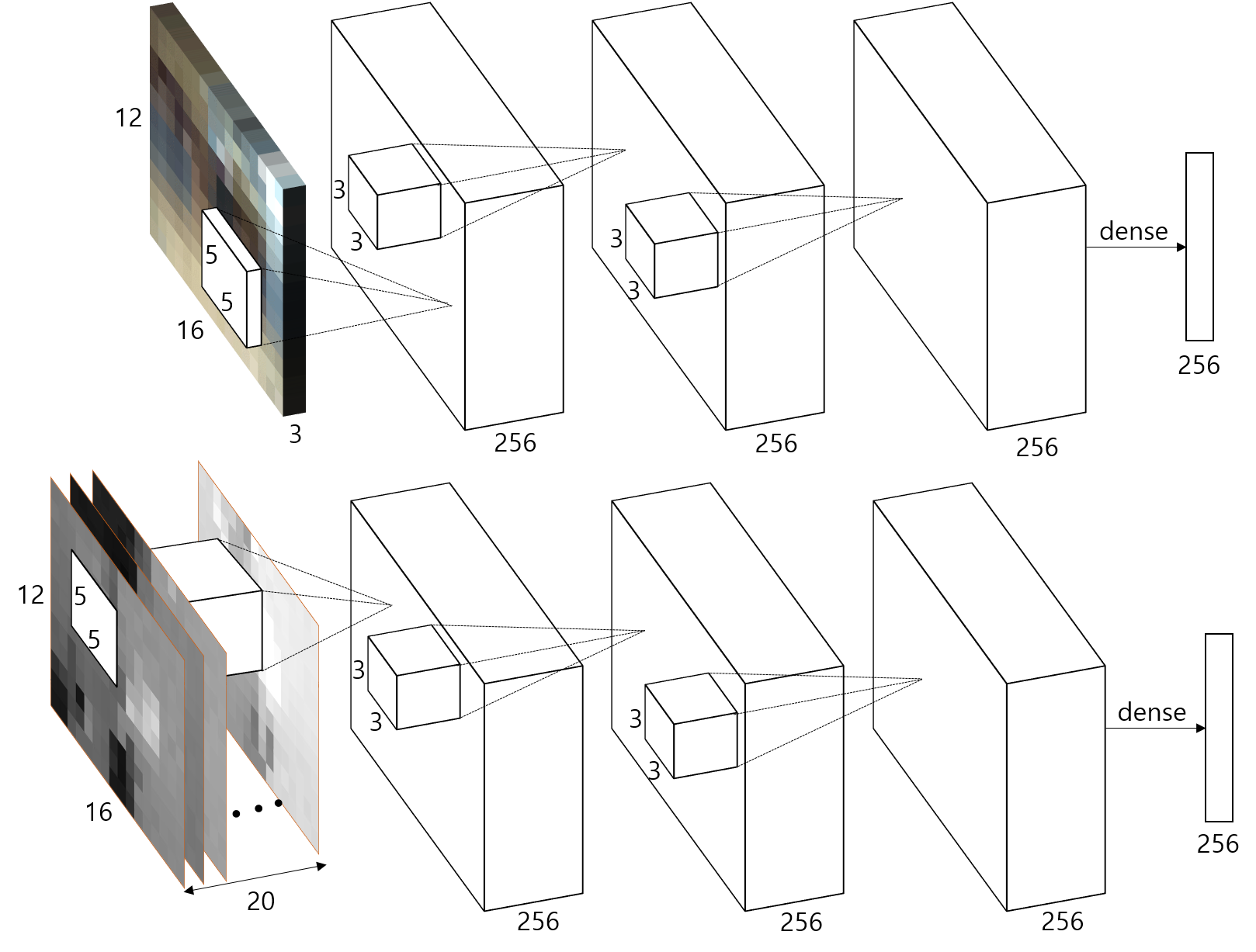}
\end{center}
   \caption{The detailed architecture used in our two-stream CNN designed for 16x12 extreme low resolution videos. This two-stream CNN is applied to each frame of the video.}
\label{fig:two-stream}		
\end{figure}

\subsection{Two-stream convolutional neural network}

We design a new two-stream convolutional neural network model for low resolution videos. Similar to other two-stream CNNs, one stream of our model takes the raw image as its input (spatial stream) and the other stream takes the concatenation optical flows (temporal stream) computed from LR images. We used 16x12 as the spatial resolution of our LR videos. More specifically, our spatial stream takes RGB pixel values of each frame as an input (i.e., the input dimensionality is 16x12x3) and the temporal stream takes 10-frame concatenation of X and Y optical flow images (i.e., 16x12x20). X and Y optical flow images are constructed by computing  ``x (and y) optical flow magnitude'' per pixel. Figure \ref{fig:two-stream} illustrates parameters used in our two-stream architecture.

We used the TV-L1 optical flow extraction algorithm \cite{zach2007duality}. More specifically, our optical flows are computed by (1) bruteforcely resizing a 16x12 video to 256x256 using a standard bicubic interpolation, (2) applying the dual TV-L1 optical flow algorithm, and (3) resizing the result back to 16x12. No HR information was used in any part of our process, since we assume only one (unlabeled) LR video is provided in the testing phase. 

\begin{figure}
\begin{center}
   \includegraphics[width=1.0\linewidth]{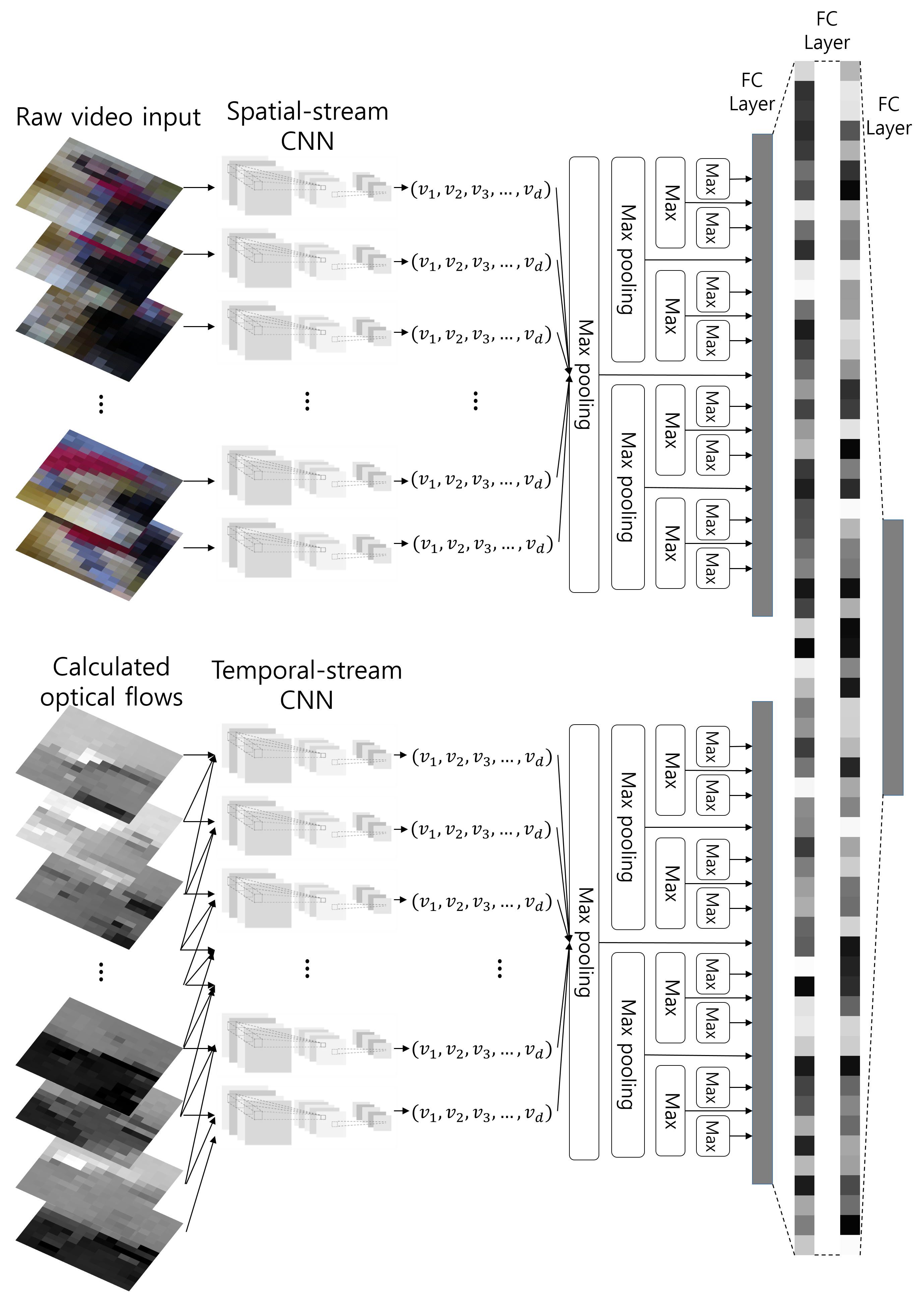}
\end{center}
   \caption{Our two-stream CNN model with temporal pyramid. This applies two-stream models from Figure \ref{fig:two-stream} for each frame, and then takes temporal max pooling with different intervals to perform the video classification.}
\label{fig:two-stream-pyramid}		
\end{figure}

Our two-stream network is applied for each frame of the video, and they are summarized using a temporal pyramid similar to \cite{ryoo15} to generate a single video representation. Let $h(V^t)$ be the two-stream network being applied to each frame $V^t$ of video $V$ at time $t$. Then, our representation $f(V ; \theta)$ is computed by 
\begin{equation}
\begin{split}
x = f(V ; \theta) = fc(&\max_{t \in [0,T]} h(V^t), \max_{[0,T/2]} h(V^t),\\
                    &\max_{[T/2,T]} h(V^t), \max_{[0,T/4]} h(V^t), \cdots)
\end{split}
\end{equation}
where $,$ denotes the vector concatenation operator, $T$ is the number of frames in the video $V$, and $fc$ denotes a set of fully connected layers to be applied on top of the concatenation. The size of $h(V^t)$ is 512-D: $256 \times 2$. $\theta$ is a set of parameters in our CNN, which we need to learn from the training data. Here, $\max$ is a temporal max pooling operator that computes the maximum of each element. In our experiments, the temporal pyramid of level 4 was used (i.e., a total of $15$ max pooling). Figure \ref{fig:two-stream-pyramid} shows the overall architecture.

Attaching more fully connected layers and a softmax layer to $f(V)$ would enable the learning of the activity video classifier. Let $g$ be such layers. Then, $y = g(f(V ; \theta))$ where $y$ is the activity class label. Training $g(f(V ; \theta))$ with the classification loss using low resolution videos generated using transforms will provide us the basic video classification model.

\subsection{Multi-Siamese CNN}
\label{subsec:multi-siamese}

Although the above two-stream network design is able to classify activity videos by learning model parameters optimized for the classification, it does not consider the property of extreme low resolution videos that different transforms applied to the same scene result different LR data. In order for the classifier to better take advantage of such nature, we require the learning of the embedding space that maps different LR videos with the same semantic content to the same embedding location whatever their transforms are. This embedding (i.e., representation) learning enables training of more generalized (i.e., less overfitted) classifier, jointly optimized for both the embedding and the classification using the learned embedding in an end-to-end fashion.


\begin{figure*}
\begin{center}
   \includegraphics[width=0.8\linewidth]{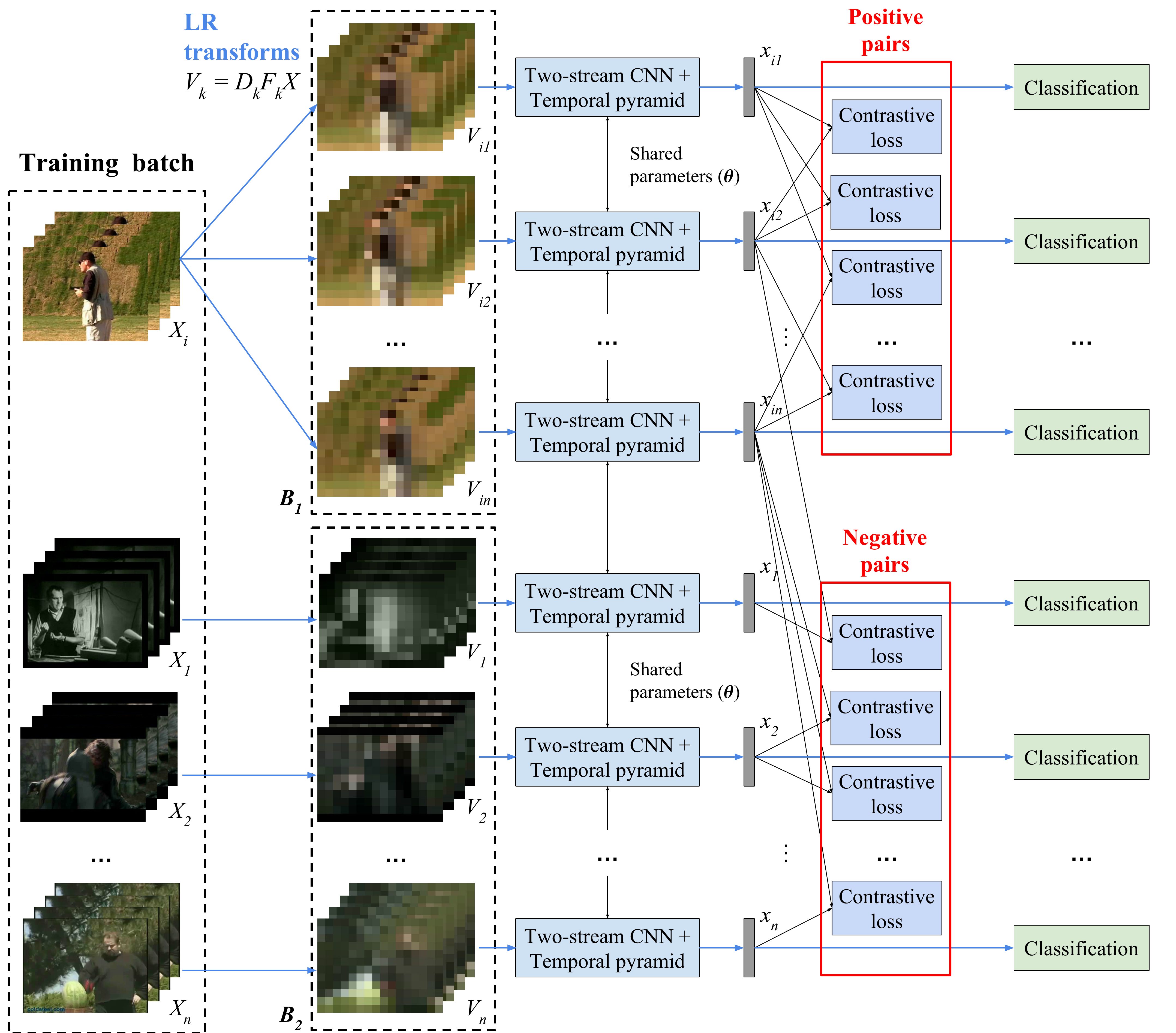}
\end{center}
   \caption{`Training' process of our multi-Siamese CNNs. It takes advantage of both contrastive and classification losses. It has $2 \cdot n$ branches sharing the parameters for the embedding and the classifier learning. In the actual testing phase, we only take advantage of one branch, applying it to each unknown low resolution test video for the classification.}
\label{fig:multi-siamese}		
\end{figure*}

{\flushleft\textbf{Siamese CNN:} A Siamese neural network is the concept of having two networks sharing the same parameters, often used to learn the similarity measure between two inputs \cite{hadsell2006dimensionality,bell15siggraph}. The objective of a Siamese network (with a contrastive loss function) is to learn the embedding space that places similar items (i.e., LR videos in our case) nearby. More specifically, it is trained with positive and negative pairs of items as training examples, where a positive pair corresponds to samples that need to stay close in the embedding space and a negative pair corresponds to samples that need to stay far away.}

Let $x = f(V ; \theta)$ be our CNN. Then, during the training, we are obtaining $x_i = f(V_i ; \theta)$ and $x_j = f(V_j ; \theta)$ by applying the same copies of the network $f(V ; \theta)$ twice to any LR video $V_i$ and $V_j$, where $(x_i, x_j)$ can either be a positive pair or a negative pair. The contrastive loss to learn the network parameters $\theta$ is described as below:
\begin{equation}
\begin{split}
    L_{siam}(\theta)=\sum_{(i,j)}^B &y'_{(i,j)} ||x_i - x_j||_2^2 + \\
                                &(1-y'_{(i,j)}) \max(0, m-||x_i - x_j||_2)^2
\end{split}
\end{equation}
where $m$ is a predetermined margin, $B$ is the batch of LR training examples being used, and $i$ and $j$ are the indexes of training pairs in the batch. $y'_{(i,j)}$ is a binary variable, which is 1 for positive pairs and 0 for negative pairs.

In our LR recognition embedding learning, a positive pair is composed of two LR videos originated from the same HR video, and a negative pair is composed of any two LR videos from different HR videos. Furthermore, since our objective is to finally classify LR videos by learning $y = g(f(V; \theta))$, we need to train the network with the combined loss function as below:
\begin{equation}
\label{eq:combined-loss}
L(\theta)= \lambda_1 L_{siam}(\theta) + \lambda_2 L_{class}(\theta)
\end{equation}
where $L_{class}(\theta)$ is the standard classification loss of the network $y = g(f(V ; \theta))$, and $\lambda_1$ and $\lambda_2$ are the weights.


{\flushleft\textbf{Multi-Siamese CNN:} Different from the standard Siamese network that only has two copies (i.e., branches) of the network sharing parameters, we designed a new model with $2 \cdot n$ network copies sharing the same parameters $\theta$ for $f(V ; \theta)$. The idea is to make each copy correspond to each of the $n$ different LR transformations (i.e., $F_k$), so that we can enforce their embedding distance to be small using a contrastive loss. In addition, we have $n$ more copies of the network to form negative training pairs by using videos not corresponding to the scene of the first $n$ branches. Figure \ref{fig:multi-siamese} illustrates our network.}

Let $x_{ik} = f(V_{ik} ; \theta)$, where $V_{ik}$ is obtained by applying the transform $F_k$ to $X_i$. Based on the batch $B$ of `original HR training videos', we randomly prepare two types of batches: $B_1$ is a batch of LR videos generated from a single HR video $X_i$, and $B_2$ is a batch with randomly selected LR videos. We use $B_1$ to generate positive pairs, and $B_1$ and $B_2$ to generate negative pairs. The sizes of $B_1$ and $B_2$ have to be $n$. For each example $X_i$ in $B$, we apply $n$ different LR transforms to get $B_1$, and provide each of the resulting $V_{ik} = D_k F_k X_i$ to the first $n$ branches of our multi-Siamese network. The LR examples $V_j$ in $B_2$ are provided to the remaining $n$ branches of the Siamese network directly. Our new loss function is formulated as:
\begin{equation}
\begin{split}
L_{multi}(\theta) = \sum_{i \in B} &\Bigg[ \sum_{(k, l) \in B_1} ||x_{ik} - x_{il}||_2^2 + \max(0, \\
        &~~n^2 \cdot m^2 - ( \sum_{k}\sum_{j \in B_2} ||x_{ik} - x_j||_2^2 ))\Bigg]
\end{split}
\end{equation}
That is, in our model, we consider multiple LR transforms simultaneously for the embedding learning. The new loss function essentially takes every pair of $n$ LR transforms as positive pairs, and also considers the same number of negative pairs using a separate batch.

The final loss function is computed by combining the above multi-Siamese contrastive loss and the standard classification loss as done in Equation \ref{eq:combined-loss}: $L(\theta)= \lambda_1 L_{multi}(\theta) + \lambda_2 \sum L_{class}(\theta)$. The overall process of our multi-Siamese embedding and classifier learning is summarized in Figure \ref{fig:multi-siamese}. This can be more specifically viewed as a Siamese CNN with multiple contrastive loss (from different LR pairs) combined. It is generalizing and extending the Siamese embedding learning beyond triplets by explicitly considering the multi-pairing of LR transforms.

We used three fully connected layers for the embedding learning and the classification. After the temporal pyramid, we obtain an intermediate representation of 7680-D per video (i.e., $15 \times 256 \times 2$). We then have the two fully connected layers with size 8192. Our embedding learning was done after this 2nd fully connected layer, making our $x$ to have the dimensionality of 8192-D. The classification was performed by having one more fully connected layer and one soft max layer on top of that.

Notice that our model relies on the multi-Siamese contrastive loss only during the `training' process. Once trained (i.e., once the embedding space is learned), in the testing phase, it is a standard feedforward convolutional neural network. It takes exactly the same amount of computation time compared to the baseline (i.e., two-stream temporal pyramid CNN) model to classify an unknown video segment.

\section{Experiments}

\subsection{Dataset and setting}

{\flushleft\textbf{16x12 HMDB dataset:} HMDB dataset \cite{hmdb} is one of the most widely used public video datasets containing more than 7000 videos with 51 different action classes. The dataset is composed of the videos mostly collected from YouTube, including movie scenes. It often serves as a standard benchmark for the evaluation of activity classification. HMDB dataset was also used in \cite{ryoo17privacy} and \cite{chen17} for the extreme low resolution recognition evaluation. We used the HMDB dataset to allow directly comparison between our approach and those previous works.}

We resized the HMDB videos to 16x12 using the average downsampling, while also including the lens blur term and the Gaussian noise term. For the videos with non-4:3 asepct ratio, a center cropping was used. The standard evaluation setting of the dataset using 3 provided training/testing splits was followed, performing the 51-class video classification.

{\flushleft\textbf{16x12 DogCentric dataset:} DogCentric dataset \cite{dogcentric} is a smaller scale dataset (compared to HMDB), consisting of more than 200 videos with 10 different activity classes. The videos in the dataset are taken from a wearable camera, mounted on top of dogs. Such videos, taken from the actor's own viewpoint, are often called first-person videos or egocentric videos. We use this dataset to test the ability of our approach to reliably recognize activities from LR videos taken with wearable cameras. This dataset was also used in \cite{ryoo17privacy}  as their main dataset for the evaluation. Identical to the HMDB dataset case, we resized the videos to 16x12 for its testing. We followed  the standard evaluation setting of the dataset, using 10 random half-training/half-testing splits.}

{\flushleft\textbf{Hardcore Henry movie:} We newly annotated events in a first-person movie called ``Hardcore Henry (2015)'', and obtained 16x12 videos from them. It is an action movie entirely taken with first-person wearable cameras. The idea was to evaluate whether we can recognize surveillance-type actions (e.g., violence) from a wearable camera where privacy-protection is most necessary. Action durations are around 3 seconds, and the task was to do binary classification of each unknown video segment (i.e., whether the segment corresponds to the action or not). A total of 67 `threat' event segments (e.g., the person getting hit, falling, ...) and 687 other segments (i.e., negative samples like `running') were annotated, and they were used for the evaluation. This is a relatively easier dataset compared to HMDB or DogCentric, in the aspect that clear camera motion caused by the event (i.e., the camera falling) is very visible even in LR.}

\begin{figure}
\begin{center}
   \includegraphics[width=1.0\linewidth]{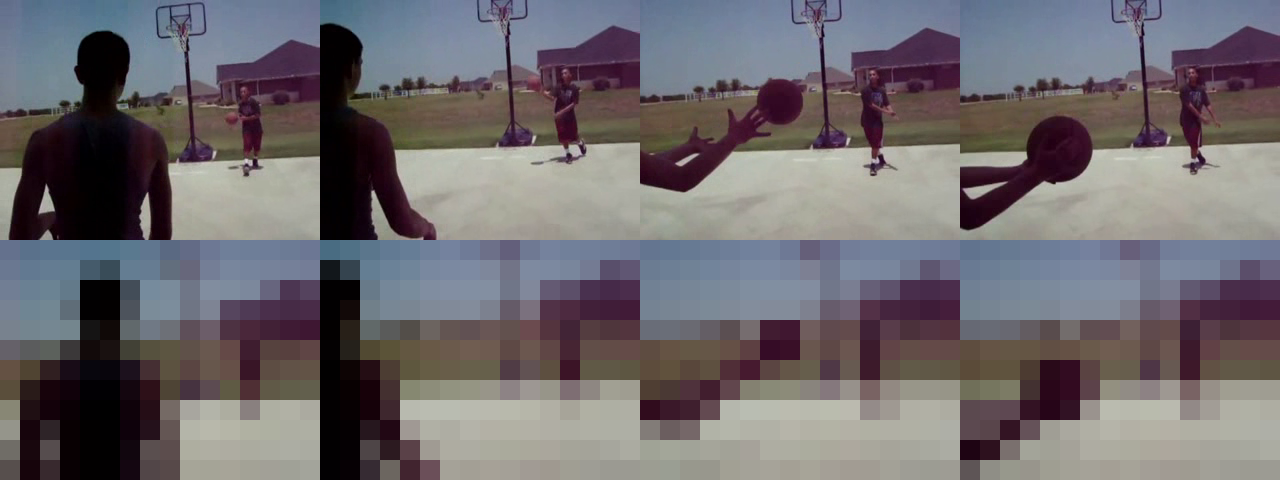}
   \includegraphics[width=1.0\linewidth]{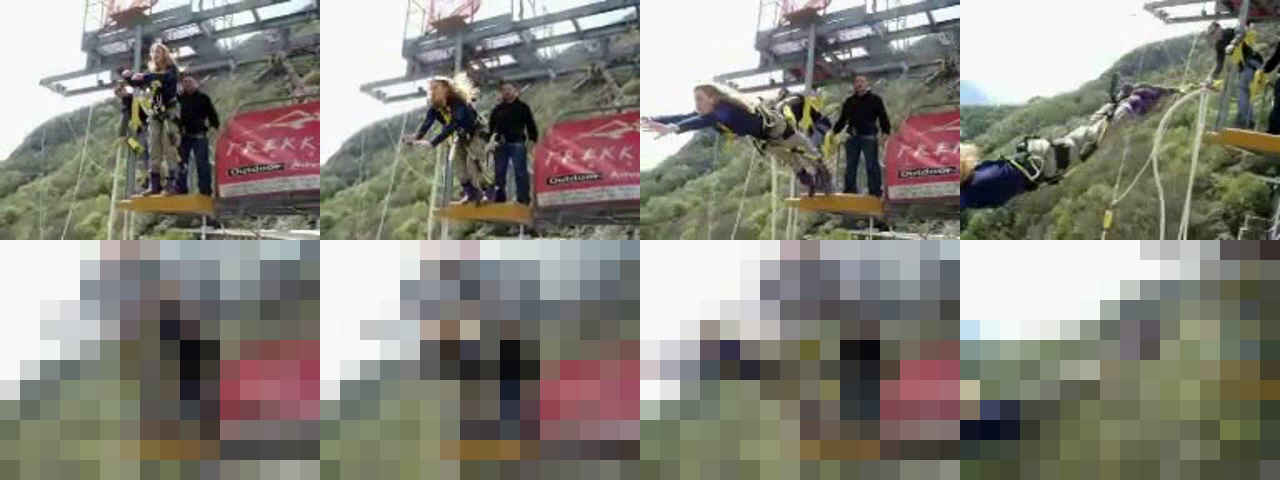}
   \includegraphics[width=1.0\linewidth]{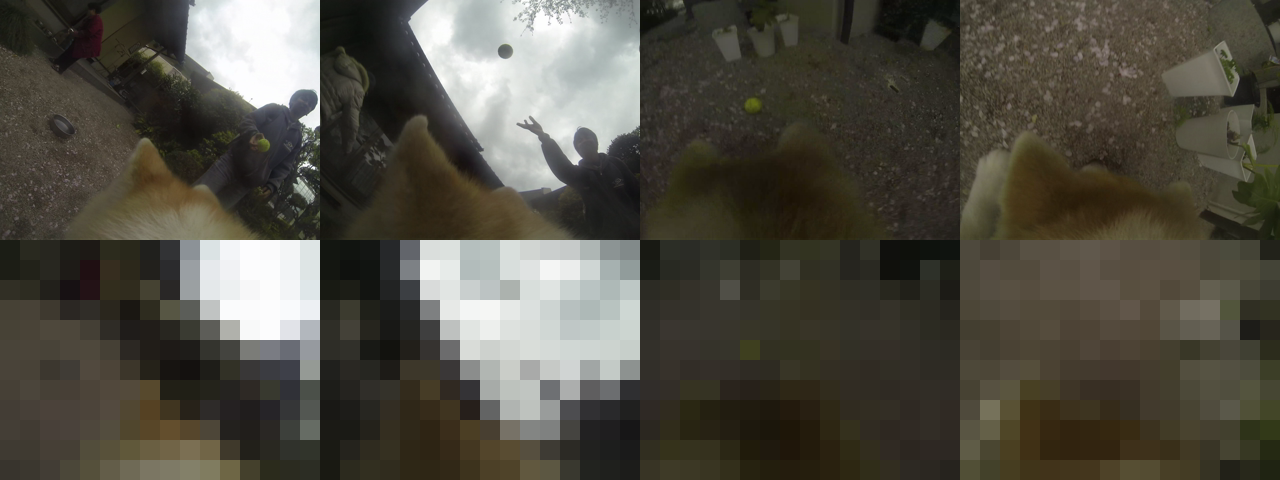}
   \includegraphics[width=1.0\linewidth]{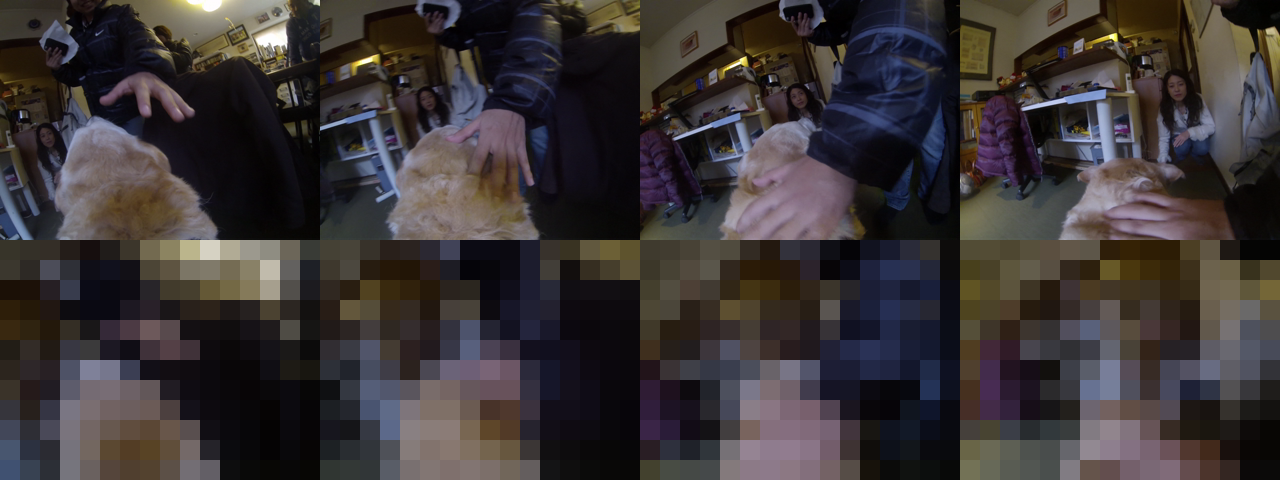}
\end{center}
   \caption{Example videos of the HMDB and DogCentric datasets. The upper rows show the original HR videos and the lower rows show the 16x12 extreme low resolution videos we use in our experiments. The first two videos are from HMDB and the other two videos are from DogCentric.}
\label{fig:videos}		
\end{figure}

Figure \ref{fig:videos} shows examples of our extreme LR videos.

\subsection{Baselines}

In addition to the previous works we are comparing our proposed approach against \cite{ryoo17privacy,chen17}, we implemented several baselines. We implemented (1) the basic one-stream CNN only taking advantage of RGB pixels values of the frame and (2) our two-stream CNN. Using these two CNNs as base components, on top of them, we evaluated three different learning approaches: We tested (i) learning these models without using multiple LR transforms (i.e., only one LR transform per training video was used). We also tested (ii) learning the models with multiple LR transforms but without the embedding learning, to compare them against (iii) our approach of using Siamese embedding learning described in the previous section. As a result, a total of 2x3 methods were tested.

The approach (ii) can be viewed as a standard data augmentation (DA) method commonly used in previous works (e.g., \cite{karpathy14}), using the exact same set of LR training videos as our approach (iii) is using. The comparison between (ii) and (iii) will confirm the benefit of our approach.


\subsection{Training}

The baselines and our approaches used the same amount of training videos provided in each dataset setting.

There were two stages in our learning process. In the first stage, we trained the two streams of our network separately using per-frame labels. The spatial stream of our two-stream network (taking a RGB frame as an input) was pre-trained using the ImageNet dataset on object classification task. The temporal stream was trained directly based on optical flows from HMDB video frames. Once such first-stage training is done, in the second stage, our entire model with the Siamese CNN architecture and the attached classifier is jointly trained. Both the activity classification loss and the contrastive loss were used to train the model in our approach.

The number of LR transforms we used in our experiments (i.e., $n$) was 75. We considered the translation of $\{-5, -2.5, 0, +2.5, +5\} \%$ in X direction and of $\{-5, 0, +5\} \%$ in Y direction, providing us a total of 15 motion transforms $F_k$. In addition, we have three different rotations with the angle $\{-10, -5, 0, 5, 10\}$ degrees, giving us a total of 75 transforms. These 75 transforms were used as our $S = \{F_k\}^{75}_{k=1}$.

For the training of the models, a standard early stopping strategy using validation errors was used to check the convergence, avoiding overfitting. Because of the fact that there is randomness in the CNN training, we repeated our experiments for 10 times and are reporting the mean and standard deviations.

\subsection{Evaluation}

We first conducted experiments with the HMDB dataset resized to 16x12, measuring 51-class classification accuracies. A total of six methods mentioned above (i.e., 5 baselines and our approach) were first compared. Table \ref{table:hmdb} illustrates the accuracies obtained by these six methods. We are able to observe that our proposed LR two-stream CNN performs a lot better than the single-stream version of the same approach. Furthermore, we can confirm that our concept of using multiple LR transforms and learning the embedding space using our `multi-Siamese architecture' is meaningfully benefiting the overall classification of the activities.

Although the `data augmentation' and our `multi-Siamese' method take advantage of the exact same amount of LR training videos, our method obtained superior results. Our multi-Siamese uses the contrastive loss to explicitly benefit from the knowledge that ``intermediate representations caused by different LR transformations should stay similar'', thereby learning transform-invariant embedding space. This allows the learning of the classifier more robust to transforms and less overfitted to the training data.

\begin{table}
	\caption{Classification accuracies (\%) measured with the 16x12 HMDB dataset. We report the performances of these different approaches obtained from multiple training epochs with the standard early stopping strategy. We are reporting the mean and standard deviation of each method.}
	\label{table:hmdb}

	\center
	\setlength\extrarowheight{0.5pt}

		\begin{tabular}	{c|c|c}
			\hline 	Approach & One-Stream & Two-Stream \tabularnewline
			\hline 	Baseline CNN & 25.08 $\pm$ 0.40	& 	31.50 $\pm$ 0.30  \tabularnewline
			        Data augmentation & 25.17 $\pm$ 0.24	& 35.34 $\pm$ 0.41  \tabularnewline
			        Our multi-Siamese & 26.21 $\pm$ 0.27	& 	\textbf{37.70} $\pm$ 0.17	  \tabularnewline
			\hline
		\end{tabular}

\end{table}

\begin{table}
	\caption{A table comparing our approach with previous state-of-the-arts on the \textbf{16x12} HMDB dataset. Note that \cite{chen17} is the two-stream version of \cite{lrface16}, extending it for the video recognition.}
	\label{table:hmdb-comp}

	\center
	\setlength\extrarowheight{0.5pt}

		\begin{tabular}	{c|c}
			\hline 	Approach & Accuracy \tabularnewline
			\hline 	
			        3-layer CNN \cite{ryoo17privacy} & 20.81 \%	 \tabularnewline
			        ResNet-32 \cite{resnet2016} & 22.37 \%	 \tabularnewline
			        PoT \cite{ryoo15} & 26.57 \% \tabularnewline
			        ISR \cite{ryoo17privacy} & 28.68 \%	 \tabularnewline
			        \cite{chen17} & 29.2 \% \tabularnewline
			\hline
			        Our two-stream CNN with pyramid  & 31.50	\%  \tabularnewline
			        Ours    & \textbf{37.70} \%	  \tabularnewline
			\hline
		\end{tabular}

\end{table}

In Table \ref{table:hmdb-comp}, we compare our approach with the reported results of the state-of-the-arts. In addition to the reported performances, we also tested the ResNet with 32 layers \cite{resnet2016}. The ResNet was pre-trained with 16x12 ImageNet and fine-tuned with 16x12 HMDB frames. We are able to clearly confirm that our proposed approach significantly outperforms the recent previous works, with more than +8\% gap. Our approach with the embedding learning using the two-stream multi-Siamese CNN obtained the best known result on the 16x12 activity recognition. Our approach was particularly effective for HMDB videos, since humans appearing in the videos are very small, causing LR videos to have very different pixel values per transform. Our approach captures such properties using the multi-Siamese embedding learning, thus obtaining a much superior performance.

In addition, we conducted the same set of experiments with the DogCentric activity dataset. Five baseline approaches were compared against our approach in Table \ref{table:dog} as it was done with HMDB, and Table \ref{table:dog-comp} shows classification accuracies of the state-of-the-art extreme low resolution activity recognition approaches compared with ours. We confirm once more that our approach obtains the best accuracy on this low resolution activity recognition task. 

Finally, we checked our method's ability to perform binary event detection given segments from continuous videos using the Hardcore Henry dataset. We measured the precision and recall values of detecting the `threat' event. F1-scores based on the precision and recall are measured. The results were: baseline 0.838 vs. data augmentation 0.871 vs. our multi-Siamese 0.885.

Our approach runs in real-time ($\sim$50 fps) on a Nvidia Jetson TX2 mobile GPU card with the TensorFlow library, when the Farneback algorithm is used for optical flows.

\begin{table}
	\caption{Classification accuracies (\%) measured with the 16x12 DogCentric dataset. We report the average performance of the approaches.}
	\label{table:dog}

	\center
	\setlength\extrarowheight{0.5pt}

		\begin{tabular}	{c|c|c}
			\hline 	Approach & One-Stream & Two-Stream \tabularnewline
			\hline 	Baseline CNN &  53.05	& 	61.25   \tabularnewline
			        Data augmentation & 57.61		& 68.09  \tabularnewline
			        Our multi-Siamese & 59.08	& 	\textbf{69.43}	  \tabularnewline
			\hline
		\end{tabular}
\end{table}

\begin{table}
	\caption{Comparing our approach with previous state-of-the-art results reported on the \textbf{16x12} DogCentric activity dataset. \cite{wang13} performed poorly since no trajectories were extracted from 16x12.}
	\label{table:dog-comp}

	\center
	\setlength\extrarowheight{0.5pt}

		\begin{tabular}	{c|c}
			\hline 	Approach & Accuracy \tabularnewline
			\hline 	
			        Iwashita et al.  \cite{dogcentric} & 46.2 \%	 \tabularnewline
			        ITF \cite{wang13} & 10.0 \% \tabularnewline
			        PoT \cite{ryoo15} & 64.6 \% \tabularnewline
			        ISR \cite{ryoo17privacy} & 67.36 \%	 \tabularnewline
			\hline
			        Our two-stream CNN with pyramid  & 61.25	\%  \tabularnewline
			        Ours    & \textbf{69.43} \%	  \tabularnewline
			\hline
		\end{tabular}

\end{table}

\section{Conclusion}

We presented a new approach for human activity recognition from extreme low resolution videos. A new two-stream Siamese convolutional neural networks was designed for the low resolution videos. The idea was to explicitly capture the inherent property of LR videos that two images originated from the exact same scene often have totally different pixel (i.e., RGB) values depending on their LR transformations. Our approach learns the shared embedding space that maps LR videos with the same content to the same location regardless of their transformations, while jointly optimizing it for the classification. Our experimental results confirmed that the proposed method outperforms all previous works by a meaningful margin.

\section*{Acknowledgement}

This research was conducted as a part of EgoVid Inc.'s research activity on privacy-preserving computer vision. This research was supported by the Tech Incubator Program for Startup Korea (TIPS), ``deep learning-based low resolution video analysis,'' and the Miraeholdings grant funded by the Korean government (Ministry of Science and ICT). Yang is the corresponding author.

\bibliographystyle{aaai}
\bibliography{low-resolution}

\begin{thebibliography}{}

\bibitem[\protect\citeauthoryear{Aggarwal and Ryoo}{2011}]{ryoo-review}
Aggarwal, J.~K., and Ryoo, M.~S.
\newblock 2011.
\newblock Human activity analysis: A review.
\newblock {\em ACM Computing Surveys} 43:16:1--16:43.

\bibitem[\protect\citeauthoryear{Bell and Bala}{2015}]{bell15siggraph}
Bell, S., and Bala, K.
\newblock 2015.
\newblock Learning visual similarity for product design with convolutional
  neural networks.

\bibitem[\protect\citeauthoryear{Chen \bgroup et al\mbox.\egroup
  }{2017}]{chen17}
Chen, J.; Wu, J.; Konrad, J.; and Ishwar, P.
\newblock 2017.
\newblock Semi-coupled two-stream fusion convnets for action recognition at
  extremely low resolutions.
\newblock In {\em WACV}.

\bibitem[\protect\citeauthoryear{Cheng \bgroup et al\mbox.\egroup
  }{2017}]{cheng17emotion}
Cheng, B.; Wang, Z.; Zhang, Z.; Li, Z.; Liu, D.; Yang, J.; Huang, S.; and
  Huang, T.~S.
\newblock 2017.
\newblock Robust emotion recognition from low quality and low bit rate video:
  {A} deep learning approach.
\newblock In {\em ACII}.

\bibitem[\protect\citeauthoryear{Dai \bgroup et al\mbox.\egroup }{2015}]{dai15}
Dai, J.; Saghafi, B.; Wu, J.; Konrad, J.; and Ishwar, P.
\newblock 2015.
\newblock Towards privacy-preserving recognition of human activities.
\newblock In {\em ICIP}.

\bibitem[\protect\citeauthoryear{Efros \bgroup et al\mbox.\egroup
  }{2003}]{efros2003recognizing}
Efros, A.~A.; Berg, A.~C.; Mori, G.; and Malik, J.
\newblock 2003.
\newblock Recognizing action at a distance.
\newblock In {\em ICCV}.

\bibitem[\protect\citeauthoryear{Hadsell, Chopra, and
  LeCun}{2006}]{hadsell2006dimensionality}
Hadsell, R.; Chopra, S.; and LeCun, Y.
\newblock 2006.
\newblock Dimensionality reduction by learning an invariant mapping.
\newblock In {\em CVPR}.

\bibitem[\protect\citeauthoryear{He \bgroup et al\mbox.\egroup
  }{2016}]{resnet2016}
He, K.; Zhang, X.; Ren, S.; and Sun, J.
\newblock 2016.
\newblock Deep residual learning for image recognition.
\newblock In {\em CVPR}.

\bibitem[\protect\citeauthoryear{Iwashita \bgroup et al\mbox.\egroup
  }{2014}]{dogcentric}
Iwashita, Y.; Takamine, A.; Kurazume, R.; and Ryoo, M.~S.
\newblock 2014.
\newblock First-person animal activity recognition from egocentric videos.
\newblock In {\em ICPR}.

\bibitem[\protect\citeauthoryear{Karpathy \bgroup et al\mbox.\egroup
  }{2014}]{karpathy14}
Karpathy, A.; Toderici, G.; Shetty, S.; Leung, T.; Sukthankar, R.; and Fei-Fei,
  L.
\newblock 2014.
\newblock Large-scale video classification with convolutional neural networks.
\newblock In {\em CVPR}.

\bibitem[\protect\citeauthoryear{Kuehne \bgroup et al\mbox.\egroup
  }{2011}]{hmdb}
Kuehne, H.; Jhuang, H.; Garrote, E.; Poggio, T.; and Serre, T.
\newblock 2011.
\newblock {HMDB}: a large video database for human motion recognition.
\newblock In {\em ICCV}.

\bibitem[\protect\citeauthoryear{Ng \bgroup et al\mbox.\egroup
  }{2015}]{google15}
Ng, J.~Y.; Hausknecht, M.~J.; Vijayanarasimhan, S.; Vinyals, O.; Monga, R.; and
  Toderici, G.
\newblock 2015.
\newblock Beyond short snippets: Deep networks for video classification.
\newblock In {\em CVPR}.

\bibitem[\protect\citeauthoryear{Piergiovanni, Fan, and
  Ryoo}{2017}]{piergiovanni2016learning}
Piergiovanni, A.; Fan, C.; and Ryoo, M.~S.
\newblock 2017.
\newblock Learning latent sub-events in activity videos using temporal
  attention filters.
\newblock In {\em AAAI}.

\bibitem[\protect\citeauthoryear{Reddy \bgroup et al\mbox.\egroup
  }{2012}]{reddy2012human}
Reddy, K.~K.; Cuntoor, N.; Perera, A.; and Hoogs, A.
\newblock 2012.
\newblock Human action recognition in large-scale datasets using histogram of
  spatiotemporal gradients.
\newblock In {\em IEEE International Conference on Advanced Video and
  Signal-Based Surveillance (AVSS)},  106--111.

\bibitem[\protect\citeauthoryear{Ryoo \bgroup et al\mbox.\egroup
  }{2017}]{ryoo17privacy}
Ryoo, M.~S.; Rothrock, B.; Fleming, C.; and Yang, H.~J.
\newblock 2017.
\newblock Privacy-preserving human activity recognition from extreme low
  resolution.
\newblock In {\em AAAI}.

\bibitem[\protect\citeauthoryear{Ryoo, Rothrock, and Matthies}{2015}]{ryoo15}
Ryoo, M.~S.; Rothrock, B.; and Matthies, L.
\newblock 2015.
\newblock Pooled motion features for first-person videos.
\newblock In {\em CVPR}.

\bibitem[\protect\citeauthoryear{Simonyan and Zisserman}{2014}]{simonyan14}
Simonyan, K., and Zisserman, A.
\newblock 2014.
\newblock Two-stream convolutional networks for action recognition in videos.
\newblock In {\em NIPS}.

\bibitem[\protect\citeauthoryear{Tran \bgroup et al\mbox.\egroup }{2015}]{c3d}
Tran, D.; Bourdev, L.; Fergus, R.; Torresani, L.; and Paluri, M.
\newblock 2015.
\newblock Learning spatiotemporal features with 3{D} convolutional networks.
\newblock In {\em ICCV}.

\bibitem[\protect\citeauthoryear{Varol, Laptev, and Schmid}{2016}]{varol16}
Varol, G.; Laptev, I.; and Schmid, C.
\newblock 2016.
\newblock Long-term temporal convolutions for action recognition.
\newblock {\em arXiv:1604.04494}.

\bibitem[\protect\citeauthoryear{Wang and Gupta}{2015}]{wang2015unsupervised}
Wang, X., and Gupta, A.
\newblock 2015.
\newblock Unsupervised learning of visual representations using videos.
\newblock In {\em CVPR},  2794--2802.

\bibitem[\protect\citeauthoryear{Wang and Schmid}{2013}]{wang13}
Wang, H., and Schmid, C.
\newblock 2013.
\newblock Action recognition with improved trajectories.
\newblock In {\em ICCV}.

\bibitem[\protect\citeauthoryear{Wang \bgroup et al\mbox.\egroup
  }{2016}]{lrface16}
Wang, Z.; Chang, S.; Yang, Y.; Liu, D.; and Huang, T.~S.
\newblock 2016.
\newblock Studying very low resolution recognition using deep networks.
\newblock In {\em CVPR}.

\bibitem[\protect\citeauthoryear{Yang and Huang}{2010}]{huang10}
Yang, J., and Huang, T.
\newblock 2010.
\newblock Image super-resolution: historical overview and future challenges.
\newblock In Milanfar, P., ed., {\em Super-resoluton imaging}. CRC Press.

\bibitem[\protect\citeauthoryear{Yeung \bgroup et al\mbox.\egroup
  }{2016}]{yeong16}
Yeung, S.; Russakovsky, O.; Mori, G.; and Fei-Fei, L.
\newblock 2016.
\newblock End-to-end learning of action detection from frame glimpses in
  videos.
\newblock In {\em CVPR}.

\bibitem[\protect\citeauthoryear{Zach, Pock, and
  Bischof}{2007}]{zach2007duality}
Zach, C.; Pock, T.; and Bischof, H.
\newblock 2007.
\newblock A duality based approach for realtime {TV-L} 1 optical flow.
\newblock {\em Pattern Recognition}  214--223.

\end{thebibliography}

\end{document}